\documentclass[11pt,a4paper]{article}

\usepackage[utf8]{inputenc}
\usepackage[T1]{fontenc}
\usepackage{times}
\usepackage{graphicx}
\usepackage{booktabs}
\usepackage{multirow}
\usepackage{colortbl}
\usepackage{amsmath}
\usepackage{hyperref}
\usepackage{xcolor}
\usepackage{tikz}
\usepackage{pgfplots}
\usepackage[margin=2.5cm]{geometry}
\usepackage{float}
\usepackage{subcaption}
\usepackage{natbib}
\usepackage{authblk}

\pgfplotsset{compat=1.18}
\usetikzlibrary{shapes, arrows.meta, positioning, calc, patterns}

\title{\textbf{IndoBERT-Relevancy: A Context-Conditioned Relevancy\\Classifier for Indonesian Text}}

\author[1]{Muhammad Apriandito Arya Saputra}
\author[2]{Andry Alamsyah}
\author[2]{Dian Puteri Ramadhani}
\author[3]{Thomhert Suprapto Siadari}
\author[4]{Hanif Fakhrurroja}

\affil[1]{SocialX \\ \texttt{apriandito@socialx.id}}
\affil[2]{Center of Excellence SAKTI, Research Institute Intelligent Business \& Sustainable Economy, Telkom University \\ \texttt{\{andry, dianpramadhani\}@telkomuniversity.ac.id}}
\affil[3]{Biomedical Engineering Study Program, School of Electrical Engineering, Telkom University \\ \texttt{thomhert@telkomuniversity.ac.id}}
\affil[4]{Research Center for Smart Mechatronics, National Research and Innovation Agency (BRIN) \\ \texttt{hani010@brin.go.id}}

\date{}

\begin{document}

\maketitle

\begin{abstract}
Determining whether a piece of text is relevant to a given topic is a fundamental task in natural language processing, yet it remains largely unexplored for Bahasa Indonesia. Unlike sentiment analysis or named entity recognition, relevancy classification requires the model to reason about the relationship between two inputs simultaneously: a topical context and a candidate text. We introduce IndoBERT-Relevancy, a context-conditioned relevancy classifier built on IndoBERT Large (335M parameters) and trained on a novel dataset of 31,360 labeled pairs spanning 188 topics. Through an iterative, failure-driven data construction process, we demonstrate that no single data source is sufficient for robust relevancy classification, and that targeted synthetic data can effectively address specific model weaknesses. Our final model achieves an F1 score of 0.948 and accuracy of 96.5\%, handling both formal and informal Indonesian text. The model is publicly available at HuggingFace.
\end{abstract}

% ============================================================
\section{Introduction}
% ============================================================

For many languages, the building blocks of text understanding are well-established. Models for sentiment analysis, topic classification, and named entity recognition are widely available and well-studied. But there is a more fundamental question that often precedes these tasks: \emph{is this text even about the topic I care about?}

This question, which we call \emph{context-conditioned relevancy classification}, is deceptively difficult. It requires the model to take two inputs: a description of a topic (the context) and a candidate text, then determine whether the text is meaningfully related to that topic. This is not a simple keyword lookup. A text can be highly relevant without mentioning any obvious keywords, and it can contain every keyword in the dictionary while discussing something entirely different.

For Bahasa Indonesia, this task has received remarkably little attention. While Indonesian NLP has made significant progress in recent years, with benchmarks like IndoNLU \citep{wilie2020indonlu} establishing baselines for sentiment analysis, natural language inference, and question answering, no dedicated model or dataset exists for context-conditioned relevancy classification. This gap is significant because Indonesia is one of the largest producers of digital text in the world, and the need to filter, sort, and prioritize this text by topical relevance is immense.

In this paper, we set out to fill this gap. We describe the construction of IndoBERT-Relevancy, a model that takes a topic and a text as input and outputs a relevancy judgment. Along the way, we encountered a challenge that we believe is instructive beyond our specific task: a model trained on one style of text fails when confronted with another. The solution, as we will show, is not simply more data, but the \emph{right kind} of data, acquired through an iterative process of identifying failures and addressing them directly.

Our contributions are:
\begin{enumerate}
    \item A \textbf{novel dataset} of 31,360 labeled text-context pairs across 188 topics and 12 domains, constructed specifically for Indonesian relevancy classification.
    \item An \textbf{iterative, failure-driven training methodology} that progressively improves model robustness through targeted data augmentation.
    \item A \textbf{publicly available model} that achieves state-of-the-art performance on this task, available for integration into downstream applications.
\end{enumerate}

% ============================================================
\section{Task Definition}
% ============================================================

We define context-conditioned relevancy classification as follows. Given a \emph{context} $c$ (a short description of a topic, such as ``Inflation and price stability'') and a \emph{text} $t$ (a sentence or short paragraph), the task is to predict a binary label $y \in \{0, 1\}$ indicating whether $t$ is relevant to $c$.

This formulation differs from standard text classification in an important way: the model must reason about the \emph{relationship} between two inputs, not just the properties of a single input. A text like ``prices keep going up'' is relevant to ``inflation'' but not to ``volcanic eruption,'' even though the text itself has the same properties in both cases. The context is what gives the classification its meaning.

This also distinguishes our task from semantic textual similarity, which measures the degree of meaning overlap between two texts. Two texts can be semantically dissimilar yet both relevant to the same topic, and two texts can be semantically similar yet relevant to different topics. Relevancy is a pragmatic judgment, not a purely semantic one.

% ============================================================
\section{Why This Is Hard for Indonesian}
% ============================================================

Indonesian presents unique challenges for relevancy classification that go beyond what is typically encountered in English-language NLP.

\subsection{The Register Spectrum}

Bahasa Indonesia exists on a wide spectrum of formality. At one end, news articles and official documents use standard Indonesian with complete sentences and formal vocabulary. At the other end, social media posts employ heavy use of slang, abbreviations, regional expressions, and code-mixing with English and local languages. Table~\ref{tab:register} illustrates this spectrum for a single topic.

\begin{table}[H]
\centering
\caption{The same topic expressed across the register spectrum. All four texts are relevant to ``fuel price increases,'' but they share almost no vocabulary.}
\label{tab:register}
\small
\begin{tabular}{lp{7cm}}
\toprule
\textbf{Register} & \textbf{Example Text} \\
\midrule
Formal news & ``Pertamina resmi menaikkan harga Pertalite menjadi Rp10.000 per liter'' \\
Informal news & ``BBM naik lagi, Pertamina umumkan harga baru'' \\
Social media & ``gila bensin naik lagi, dompet makin tipis nih'' \\
Highly informal & ``isi full tank skrg kayak bayar cicilan motor wkwk'' \\
\bottomrule
\end{tabular}
\end{table}

A model trained only on formal text will struggle with the bottom two examples, not because they are harder in any absolute sense, but because they use an entirely different vocabulary to express the same meaning.

\subsection{Implicit Expression}

Perhaps more challenging than register variation is the phenomenon of implicit expression. Indonesian social media users frequently discuss topics through personal anecdotes, complaints, and observations without ever naming the topic explicitly. Someone complaining that ``belanja di pasar makin sedikit dapet'' (``shopping at the market, getting less and less'') is clearly discussing rising prices, but the connection requires world knowledge and pragmatic inference that goes beyond pattern matching.

This implicit expression is not a marginal phenomenon. In our analysis, a significant portion of topically relevant social media text contains no explicit topic keywords at all. Any model that relies on keyword overlap, whether explicitly through string matching or implicitly through training data that only contains keyword-rich examples, will systematically miss these texts.

% ============================================================
\section{Building the Dataset}
% ============================================================

No existing dataset addresses context-conditioned relevancy classification for Indonesian. We therefore constructed our own, through a process that turned out to be as instructive as the final model itself.

Our dataset covers 188 topical contexts organized into 12 thematic domains (Table~\ref{tab:domains}), ranging from economics and politics to health, technology, and sports. This breadth is intentional: a relevancy classifier should be general-purpose, not tied to any single domain.

\begin{table}[H]
\centering
\caption{Distribution of 188 topical contexts across 12 thematic domains.}
\label{tab:domains}
\small
\begin{tabular}{lr}
\toprule
\textbf{Domain} & \textbf{Contexts} \\
\midrule
Social \& Community Issues & 22 \\
Industry \& Business & 20 \\
Economics \& Finance & 17 \\
Health & 13 \\
Law \& Crime & 12 \\
Environment \& Disasters & 12 \\
Politics \& Governance & 10 \\
Digital \& Technology & 9 \\
Infrastructure \& Transportation & 8 \\
Energy & 7 \\
Others (Sports, Religion, Culture, International) & 58 \\
\midrule
\textbf{Total} & \textbf{188} \\
\bottomrule
\end{tabular}
\end{table}

The dataset was constructed in three stages. Crucially, each stage was motivated by specific failures observed in the model trained on the previous stage. This was not a pre-planned three-stage pipeline; it emerged naturally from the iterative process of building, testing, and improving.

\subsection{Stage 1: Formal Text Pairs}

We began by collecting article titles from Indonesian news sources across all 188 topics. For each topic, we generated candidate pairs by combining article titles with both matching and non-matching contexts, then used a large language model (GPT-4o-mini) to label each pair for relevance.

To create challenging negative examples, we employed a stratified sampling strategy based on topic similarity. Topics that are semantically close (e.g., ``inflation'' and ``food prices'') produce harder negatives than distant topics (e.g., ``inflation'' and ``badminton''), because the texts may share vocabulary while discussing genuinely different aspects.

This produced \textbf{18,798 labeled pairs}. The model trained on this data achieved an F1 of 0.860: a reasonable starting point, but with a concerning recall of only 84.5\%. One in six relevant texts was being missed.

\subsection{Stage 2: Informal Text Pairs}

Testing the Stage 1 model on informal text revealed the problem immediately. Texts like ``bensin naik lagi, dompet makin tipis'' (about fuel prices) and ``temen gw di tokped kena layoff'' (about tech layoffs) were consistently classified as not relevant, despite being obviously relevant to a human reader.

The diagnosis was clear: the model had never seen informal text during training, so it had no basis for understanding it. We addressed this by collecting real social media posts across 40 high-volume topics over an 8-week period, yielding 11,983 unique posts after deduplication. These were paired with contexts and labeled through the same LLM pipeline, with the labeling instructions explicitly noting that informal language, slang, and abbreviations should not affect the relevancy judgment.

This added \textbf{7,884 labeled pairs} to the dataset. The model retrained on the combined data improved substantially, reaching an F1 of 0.922 and recall of 91.4\%.

\subsection{Stage 3: Implicit Text Generation}

The Stage 2 model was much better at informal text, but a subtler failure pattern remained. Texts that discussed a topic \emph{without using any recognizable keywords} were still being missed. For instance, ``indomie sekarang 3500, dulu cuma 1500'' (about the rising price of instant noodles) was classified as not relevant to inflation, even though the connection is obvious to anyone familiar with Indonesian daily life.

The problem was that even our social media data, while informal in style, still tended to contain at least some topical keywords. To teach the model about implicit relevance, we needed examples that were relevant \emph{despite} the absence of keywords.

We generated these using GPT-4o-mini, with an explicit instruction: create social media text that is relevant to a given topic \emph{without using any obvious topic keywords}. The model was asked to express relevance through personal experiences, complaints, and everyday observations. We also generated hard negative examples: texts that might superficially seem related but are actually about different topics.

This added \textbf{4,678 labeled pairs}. The final combined dataset contains 31,360 pairs (Table~\ref{tab:dataset-summary}).

\begin{table}[H]
\centering
\caption{Summary of the three-stage dataset construction.}
\label{tab:dataset-summary}
\begin{tabular}{lrrrc}
\toprule
\textbf{Stage} & \textbf{Pairs} & \textbf{Relevant} & \textbf{Not Rel.} & \textbf{Contexts} \\
\midrule
1. Formal text & 18,798 & 4,584 & 14,214 & 188 \\
2. Informal text & 7,884 & 3,022 & 4,862 & 40 \\
3. Implicit text & 4,678 & 3,094 & 1,584 & 30 \\
\midrule
\textbf{Combined} & \textbf{31,360} & \textbf{10,700} & \textbf{20,660} & \textbf{188} \\
\bottomrule
\end{tabular}
\end{table}

\begin{figure}[H]
\centering
\begin{tikzpicture}[
    node distance=0.6cm,
    stage/.style={rectangle, draw=black, fill=blue!8, rounded corners, minimum width=3.8cm, minimum height=1.3cm, align=center, font=\small},
    result/.style={rectangle, draw=black!50, fill=green!8, rounded corners, minimum width=2.5cm, minimum height=0.9cm, align=center, font=\footnotesize},
    gap/.style={rectangle, draw=red!50, fill=red!5, rounded corners, minimum width=3cm, minimum height=0.7cm, align=center, font=\footnotesize\itshape},
    arrow/.style={-{Stealth[length=2.5mm]}, thick},
    gaparrow/.style={-{Stealth[length=2.5mm]}, thick, red!60, dashed},
]

% Stage 1
\node[stage] (s1) {\textbf{Stage 1}\\Formal text\\18,798 pairs};
\node[result, right=1cm of s1] (r1) {F1 = 0.860\\Recall = 84.5\%};
\node[gap, below=0.4cm of r1] (g1) {Gap: fails on\\informal text};

% Stage 2
\node[stage, below=1.8cm of s1] (s2) {\textbf{Stage 2}\\+ Informal text\\+7,884 pairs};
\node[result, right=1cm of s2] (r2) {F1 = 0.922\\Recall = 91.4\%};
\node[gap, below=0.4cm of r2] (g2) {Gap: fails on\\implicit references};

% Stage 3
\node[stage, below=1.8cm of s2] (s3) {\textbf{Stage 3}\\+ Implicit text\\+4,678 pairs};
\node[result, right=1cm of s3] (r3) {\textbf{F1 = 0.948}\\Recall = 94.8\%};

% Arrows
\draw[arrow] (s1) -- (r1);
\draw[arrow] (s2) -- (r2);
\draw[arrow] (s3) -- (r3);
\draw[gaparrow] (g1.west) -- +(-0.5,0) |- (s2.east);
\draw[gaparrow] (g2.west) -- +(-0.5,0) |- (s3.east);

\end{tikzpicture}
\caption{The iterative, failure-driven data construction process. Each stage is motivated by a specific weakness identified in the previous model. Dashed red arrows indicate the identified failure that motivated the next stage.}
\label{fig:pipeline}
\end{figure}
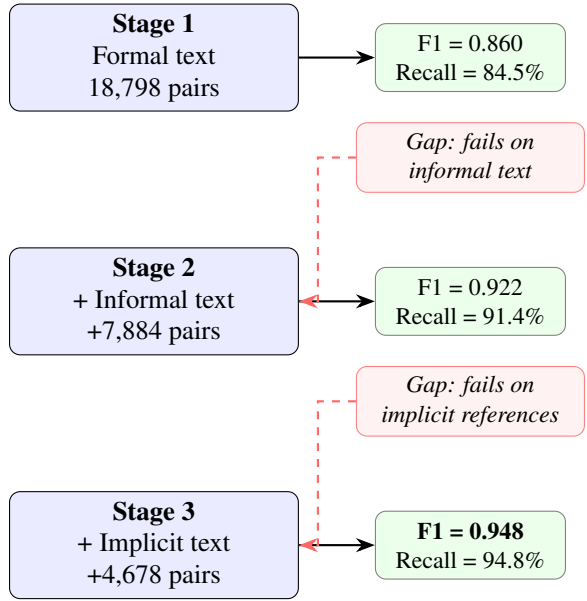

% ============================================================
\section{Model}
% ============================================================

\subsection{Architecture}

We fine-tune IndoBERT Large P2 \citep{wilie2020indonlu}, a 335-million parameter BERT model pre-trained on 23.43 GB of Indonesian text. The model takes a context-text pair as input, formatted as:

\begin{center}
\texttt{[CLS] context [SEP] text [SEP]}
\end{center}

A classification head maps the \texttt{[CLS]} representation to a binary prediction.

\subsection{Handling Class Imbalance}

Our dataset is naturally imbalanced: 34.1\% relevant and 65.9\% not relevant. Without correction, the model would learn to favor the majority class, achieving high accuracy by being overly conservative. Since missing relevant content is typically more costly than including irrelevant content, we apply inverse-frequency class weighting to the training loss. This gives approximately 2.5 times more weight to errors on relevant texts, encouraging the model to maintain high recall without sacrificing precision.

\subsection{Training}

We train for 5 epochs with a learning rate of $2 \times 10^{-5}$, batch size of 16, and maximum sequence length of 256 tokens. We use early stopping with patience of 2 epochs based on F1 score on a 15\% stratified validation set. Training was conducted on a single NVIDIA A40 GPU and completed in approximately 33 minutes.

% ============================================================
\section{Results}
% ============================================================

\subsection{Progressive Improvement}

The central result of this work is summarized in Table~\ref{tab:results}: each stage of data construction produces a measurable improvement in model performance, with the most dramatic gains in recall.

\begin{table}[H]
\centering
\caption{Model performance at each stage of data construction.}
\label{tab:results}
\begin{tabular}{llcccc}
\toprule
& \textbf{Training Data} & \textbf{Accuracy} & \textbf{Precision} & \textbf{Recall} & \textbf{F1} \\
\midrule
V1 & Formal only (18.8K) & 93.3\% & 87.8\% & 84.5\% & 0.860 \\
V2 & + Informal (26.7K) & 95.6\% & 93.0\% & 91.4\% & 0.922 \\
\textbf{V3} & \textbf{+ Implicit (31.4K)} & \textbf{96.5\%} & \textbf{94.8\%} & \textbf{94.8\%} & \textbf{0.948} \\
\bottomrule
\end{tabular}
\end{table}

\begin{figure}[H]
\centering
\begin{tikzpicture}
\begin{axis}[
    ybar,
    ylabel={Score (\%)},
    symbolic x coords={V1: Formal,V2: +Informal,V3: +Implicit},
    xtick=data,
    ymin=80,
    ymax=100,
    bar width=14pt,
    legend style={at={(0.5,-0.25)}, anchor=north, legend columns=3, font=\small},
    width=0.85\columnwidth,
    height=6cm,
    enlarge x limits=0.25,
    nodes near coords,
    nodes near coords style={font=\footnotesize},
    every node near coord/.append style={above},
]
\addplot[fill=blue!50] coordinates {(V1: Formal,87.8) (V2: +Informal,93.0) (V3: +Implicit,94.8)};
\addplot[fill=red!50] coordinates {(V1: Formal,84.5) (V2: +Informal,91.4) (V3: +Implicit,94.8)};
\addplot[fill=green!50] coordinates {(V1: Formal,86.0) (V2: +Informal,92.2) (V3: +Implicit,94.8)};
\legend{Precision, Recall, F1}
\end{axis}
\end{tikzpicture}
\caption{Model performance improves with each data source. The largest single improvement comes from adding informal text (+6.2 F1), while implicit text provides a targeted boost (+2.6 F1).}
\label{fig:results}
\end{figure}
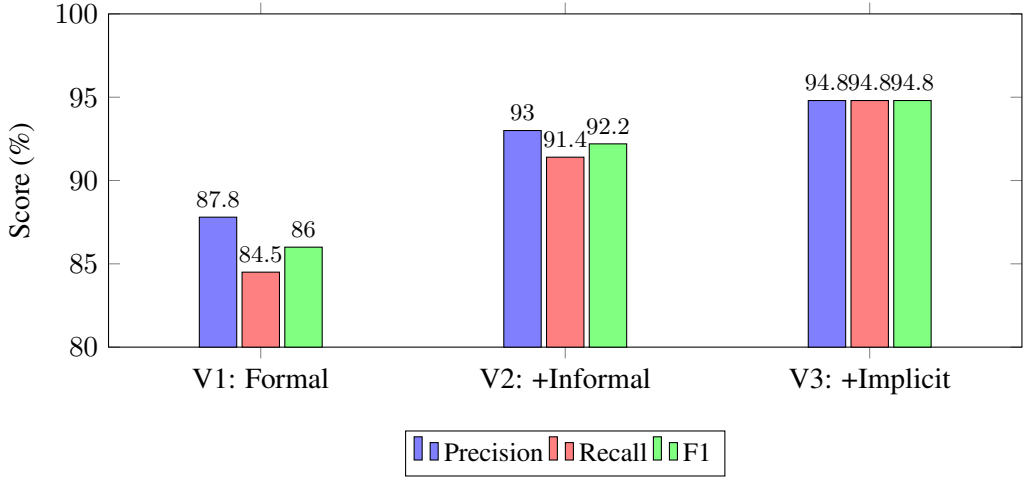

Recall improved from 84.5\% to 94.8\% across the three stages. In practical terms, this means the model went from missing approximately 1 in 6 relevant texts to missing fewer than 1 in 20.

\subsection{Validation Performance}

The final model achieves balanced performance on the validation set of 4,704 samples (Table~\ref{tab:confusion}).

\begin{table}[H]
\centering
\caption{Confusion matrix on the validation set. The near-symmetric error distribution (84 false positives vs. 83 false negatives) confirms balanced precision and recall.}
\label{tab:confusion}
\small
\begin{tabular}{ll|cc}
\toprule
& & \multicolumn{2}{c}{\textbf{Predicted}} \\
& & Not Relevant & Relevant \\
\midrule
\multirow{2}{*}{\textbf{Actual}} & Not Relevant & \cellcolor{green!15} 3,015 & \cellcolor{red!8} 84 \\
& Relevant & \cellcolor{red!8} 83 & \cellcolor{green!15} 1,522 \\
\bottomrule
\end{tabular}
\end{table}

\subsection{Training Dynamics}

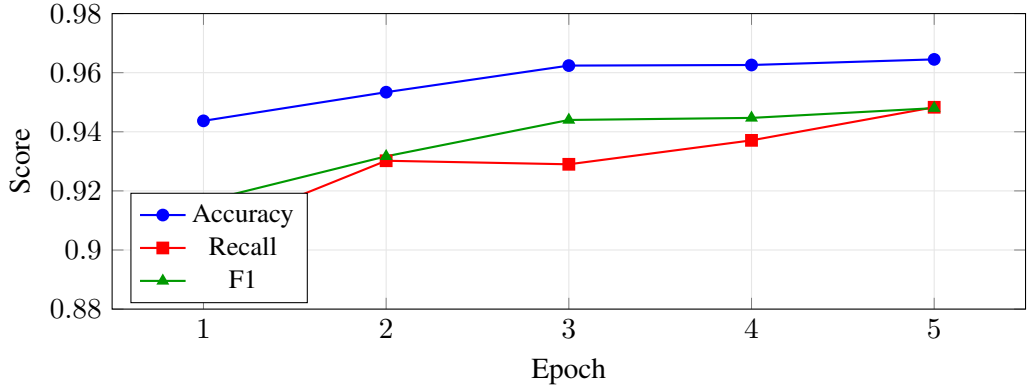
\begin{figure}[H]
\centering
\begin{tikzpicture}
\begin{axis}[
    xlabel={Epoch},
    ylabel={Score},
    xmin=0.5, xmax=5.5,
    ymin=0.88, ymax=0.98,
    xtick={1,2,3,4,5},
    legend style={at={(0.02,0.02)}, anchor=south west, font=\small},
    width=0.85\columnwidth,
    height=5.5cm,
    grid=major,
    grid style={gray!20},
]
\addplot[blue, thick, mark=*] coordinates {(1,0.9437) (2,0.9534) (3,0.9624) (4,0.9626) (5,0.9645)};
\addplot[red, thick, mark=square*] coordinates {(1,0.9053) (2,0.9302) (3,0.9290) (4,0.9371) (5,0.9483)};
\addplot[green!60!black, thick, mark=triangle*] coordinates {(1,0.9164) (2,0.9317) (3,0.9440) (4,0.9447) (5,0.9480)};
\legend{Accuracy, Recall, F1}
\end{axis}
\end{tikzpicture}
\caption{Validation metrics across training epochs for the final model (V3).}
\label{fig:training}
\end{figure}

Figure~\ref{fig:training} shows the validation metrics across training epochs. The model converges steadily, with recall being the last metric to stabilize, consistent with the class-weighted loss encouraging the model to gradually improve its sensitivity to the minority class.

\subsection{Qualitative Analysis}

Beyond aggregate metrics, we evaluated the model on hand-crafted examples designed to test its ability to handle different types of relevance expression (Table~\ref{tab:qualitative}).

\begin{table}[H]
\centering
\caption{Qualitative evaluation across text registers and relevance types. The model correctly handles explicit formal text, informal slang, and correctly rejects off-topic text.}
\label{tab:qualitative}
\small
\begin{tabular}{p{2.5cm}p{5.5cm}cr}
\toprule
\textbf{Context} & \textbf{Text} & \textbf{Pred.} & \textbf{Score} \\
\midrule
\multicolumn{4}{l}{\textit{Explicit, formal:}} \\
Monetary policy & BI tahan suku bunga acuan 6\% & Rel. & 1.000 \\
Oil prices & OPEC pangkas produksi, minyak melonjak & Rel. & 1.000 \\
\midrule
\multicolumn{4}{l}{\textit{Explicit, informal:}} \\
Fuel prices & gila bensin naik lagi, dompet makin tipis & Rel. & 1.000 \\
Tech layoffs & temen gw di tokped kena layoff semua divisi & Rel. & 0.997 \\
\midrule
\multicolumn{4}{l}{\textit{Implicit:}} \\
Food prices & harga cabai gila-gilaan, 100rb sekilo & Rel. & 0.988 \\
Corruption & duit rakyat dimakan koruptor terus & Rel. & 1.000 \\
Crypto & baru aja kena rug pull rugi 50jt & Rel. & 0.628 \\
\midrule
\multicolumn{4}{l}{\textit{Correctly rejected:}} \\
Fuel prices & makan siang apa ya hari ini & Not rel. & 0.000 \\
Elections & eh ada promo shopee 3.3 nih & Not rel. & 0.000 \\
Gambling & lagi main mobile legends rank mythic & Not rel. & 0.000 \\
\bottomrule
\end{tabular}
\end{table}

The model demonstrates robust performance across the register spectrum. Notably, it correctly identifies informal texts as relevant even when they share no vocabulary with the context description. It also correctly rejects texts that might share superficial features with the topic (e.g., ``main mobile legends'' is correctly distinguished from online gambling).

% ============================================================
\section{What We Learned}
% ============================================================

\subsection{Data Quality Over Data Quantity}

The most important lesson from this work is that the \emph{composition} of training data matters more than its \emph{volume}. Stage 3 (implicit text) comprised only 14.9\% of the final dataset, yet it produced the improvement that made the biggest qualitative difference in model behavior. A model trained on 100,000 news headlines would almost certainly perform worse than our model trained on 31,360 carefully constructed examples from three complementary sources.

\subsection{The Iterative Approach Works}

We did not design a three-stage pipeline in advance. We built a model, tested it, found where it failed, and addressed those failures. This iterative approach naturally produced a dataset that covers the space of possible inputs more effectively than any single-source collection could. We believe this methodology is broadly applicable to NLP tasks where labeled data must be constructed from scratch.

\subsection{Synthetic Data as Targeted Intervention}

The success of Stage 3 suggests a specific role for LLM-generated synthetic data: not as a replacement for real data, but as a targeted intervention to address identified model weaknesses. By constraining the generation process (``generate relevant text \emph{without} using topic keywords''), we produced examples that directly exercised the capability the model was lacking. This is fundamentally different from unconstrained data augmentation, which may produce more data without addressing the underlying gap.

\subsection{Class Weighting Is Essential}

Without class weighting, the model achieved high accuracy by being conservative, frequently classifying borderline texts as not relevant. Inverse-frequency weighting brought precision and recall into near-perfect balance (both 94.8\%), confirming that this simple technique is effective for imbalanced relevancy datasets.

% ============================================================
\section{Limitations}
% ============================================================

Several limitations should be noted. First, our model covers 188 contexts across 12 domains, but may underperform on entirely novel topics not represented in the training data. Second, some highly implicit texts remain challenging, particularly those requiring deep cultural knowledge or multi-step reasoning. Third, our synthetic data, while effective, may not fully capture the diversity and noise of real-world text. Finally, language evolves: new slang, emerging topics, and shifting discourse patterns may require periodic model updates.

% ============================================================
\section{Conclusion}
% ============================================================

We have introduced IndoBERT-Relevancy, the first dedicated context-conditioned relevancy classifier for Bahasa Indonesia. Through an iterative process of building, testing, and improving, we constructed a dataset of 31,360 labeled pairs and trained a model that achieves an F1 of 0.948 across 188 diverse topics.

The journey from an F1 of 0.860 to 0.948 was not a story of scaling up. It was a story of understanding what the model was getting wrong and finding the right data to fix it. We believe this iterative, failure-driven approach to dataset construction is a practical methodology that can be applied to many NLP tasks, particularly in languages and domains where labeled data is scarce.

The model is publicly available at: \url{https://huggingface.co/apriandito/indobert-relevancy-classifier}

\bibliographystyle{plainnat}

\end{document}